# ContextVLM: Zero-Shot and Few-Shot Context Understanding for Autonomous Driving using Vision Language Models

Shounak Sural*, Naren*, Ragunathan (Raj) Rajkumar

*Abstract*— In recent years, there has been a notable increase in the development of autonomous vehicle (AV) technologies aimed at improving safety in transportation systems. While AVs have been deployed in the real-world to some extent, a full-scale deployment requires AVs to robustly navigate through challenges like heavy rain, snow, low lighting, construction zones and GPS signal loss in tunnels. To be able to handle these specific challenges, an AV must reliably recognize the physical attributes of the environment in which it operates. In this paper, we define context recognition as the task of accurately identifying environmental attributes for an AV to appropriately deal with them. Specifically, we define 24 environmental contexts capturing a variety of weather, lighting, traffic and road conditions that an AV must be aware of. Motivated by the need to recognize environmental contexts, we create a context recognition dataset called *DrivingContexts* with more than 1.6 million context-query pairs relevant for an AV. Since traditional supervised computer vision approaches do not scale well to a variety of contexts, we propose a framework called *ContextVLM* that uses vision-language models to detect contexts using zero- and few-shot approaches. *ContextVLM* is capable of reliably detecting relevant driving contexts with an accuracy of more than $95\%$ on our dataset, while running in real-time on a 4GB Nvidia GeForce GTX 1050 Ti GPU on an AV with a latency of 10.5 ms per query.

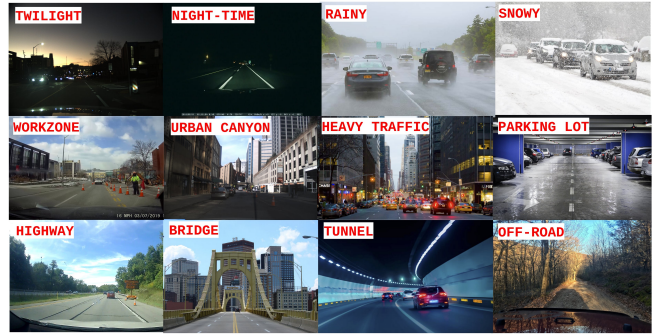

Fig. 1. Typical driving contexts that an AV has to operate in.

## I. INTRODUCTION

Autonomous vehicle (AV) deployment in the real world requires the specification and use of Operational Design Domains (ODDs). An ODD refers to road and environmental conditions in addition to geographic locations where the AV deployment has been tested extensively and is expected to operate safely. The performance of the core autonomous driving functions such as perception, planning, behavior and localization are heavily dependent on the operating environment. In particular, sensor-based perception can be affected by snow, fog, rain and low-lighting conditions. Path and speed trajectories generated by a motion planner can also benefit from operational domain knowledge such as whether the road is uphill or downhill, whether the road is paved, cobbled, gravelly or unpaved. Knowledge of workzones in the surrounding areas has significant safety implications [1]. Furthermore, whether an AV is being driven in a rural area, an urban canyon, a tunnel or a highway can significantly affect its localization performance [2], for example, because of varying Global Navigation Satellite System (GNSS) accuracy. Nevertheless, existing studies typically overlook the need for generalized context recognition that can feed and influence various layers of the AV stack.

For AVs to operate safely, to expand ODDs and enable widespread deployment, they must reliably recognize a broad range of operating conditions. A large annotated dataset is necessary to utilize and test deep-learning models. Unfortunately, many relevant driving contexts are missing in state-of-the-art AV datasets like NuScenes [3] and KITTI [4], which are biased towards city driving primarily in good weather conditions and in daytime.

Large Language Models (LLMs) and large Vision Language Models (VLMs) have gained popularity over the past few years with applications across several fields. They are particularly appealing because, unlike vision, language data is vast and highly varied to handle the long-tailed nature of many real-world tasks. Vision Language Models, in particular, are trained with millions of image-text pairs for tasks such as image captioning and visual question answering. VLMs have shown some promise towards tackling the long-tail of autonomous driving problems by modeling complex scenarios with natural language. However, to the best of our knowledge, VLMs are yet to be used for the detection of driving contexts for AVs. In this paper, we propose a practical and lightweight VLM-based approach to address the long-tailed problem of context recognition in AVs. Towards this end, we create a pipeline based on vision-language models and the *DrivingContexts* dataset for improving the performance of context recognition. Figure 1 shows a few instances from our *DrivingContexts* dataset with key driving contexts annotated. The first row shows a few adverse lighting and weather conditions, which can make perception difficult, and the next two rows show various situations that require specialized AV localization approaches and/or behavior. For widespread deployment, an AV should be able to reliably detect such contexts and make safe decisions in real-time.

*Equal Contribution.

All authors are with the Department of Electrical and Computer Engineering at Carnegie Mellon University, Pittsburgh, PA, 15213, USA. Email: {ssural,naren2,rajkumar}@andrew.cmu.edu

The key contributions of this paper are as follows.

- We identify and formalize the problem of recognizing a varied range of contexts for autonomous driving.
- We formally evaluate the applicability of large vision-language models for context detection in AVs using zero-shot and few-shot learning, circumventing the need for large, class-balanced and context-annotated datasets.
- We create a large, publicly-available dataset named *DrivingContexts* with a combination of hand-annotated and machine annnotated labels to improve VLMs for better context recognition.
- *ContextVLM* achieves an accuracy of over 95% on the *DrivingContexts* dataset and runs in real-time on an AV, demonstrating its practical usability.

The organization of the rest of the paper is as follows. Section II discusses existing literature related to this domain. Our methodology, details about our dataset, and the relevance of recognizing contexts are described in Section III. Section IV presents our experiments and an evaluation of our approach. Finally, Section V summarizes our work and identifies future directions. Our code and dataset are publicly available at https://github.com/ssuralcmu/ContextVLM.git.

## II. RELATED WORK

An AV requires safe and timely functioning of multiple subsystems including perception, localization, planning, behavior and control. An AV must also fuse the outputs of various sensors including lidars, radars, cameras, infrared sensors, GNSS sensors and IMUs appropriately. Additionally, Connected Autonomous Vehicles (CAVs) have Vehicle-to-Everything (V2X) communication subsystems that interact with other vehicles, pedestrians and traffic lights. All such sensors have different limitations and failure modes. Since various AV subsystems must handle a range of situations safely, it is necessary for an AV with modules such as perception and localization to have knowledge of the operating conditions and environmental contexts as discussed in Section I.[1]

We refer to the problem of determining the operating conditions and environmental conditions of an AV as *Context Recognition*. Prior work in this domain by Feriol et al. [5] has discussed the fusion of camera images with GNSS-based feature vectors to infer some driving contexts including canyon, open-sky, trees and urban environments. Neural networks that can infer the positions, shapes and orientations of road objects in day and nighttime have been proposed in [6]. The detection of rain in images has also been an area of interest [7]. In the context of object detection for perception, incorporating the knowledge of night-time, rainy and snowy conditions in lidar, radar and camera fusion networks can significantly improve object detection performance as well [8, 9]. In the domain of localization for autonomous vehicles, methods such as LaneMatch [2] focus on improving localization performance in GNSS-unfriendly areas such as tunnels and urban canyons with the fusion of lane-marker information from cameras. The planning and behavioral components of an AV heavily depend on the driving context, including whether the vehicle is off-road or on a paved road, or whether lane markers are available. For an AV relying on multiple approaches such as camera and lidar-based detection, such information about rainy and snowy conditions can be used for context-aware fusion of these contributing modalities. Yoneda et al. [10] have highlighted the importance of understanding contexts and failure modalities for relative sensor importance across lidar, radar and GNSS technologies. Major AV datasets such as NuScenes [3] and [4], however, lack annotations about contexts and mostly contain ideal operating conditions.

---

[1]End-to-end AI-based autonomous vehicles adopt a fundamentally different approach.

In recent years, progress has been made towards using VLMs for various tasks in the context of autonomous driving [11, 12]. End-to-end vision-language models such as DriveLM [12] and motion-forecasting methods such as MotionLM [13] have been introduced recently. VLMs can also enhance Bird's Eye View (BEV) maps for autonomous driving [14].

In contrast to the above-mentioned methods, in this paper, we focus on context understanding for autonomous vehicles in zero-shot and few-shot settings by leveraging VLMs.

## III. OUR METHODOLOGY

In this section, we first describe the problem of context recognition in detail and its importance in an AV software stack. We then introduce our *DrivingContexts* dataset followed by our approach for identifying driving contexts.

### A. Driving Contexts

Fully functional AVs must function in various road environments (e.g. urban, suburban, and highways) and different operating conditions (e.g. weather, lighting, and heavy traffic). We refer to these ambient circumstances as *Driving Contexts*. An AV may have multiple onboard sensors such as GNSS, lidar, radar, IMU, RGB/D Cameras, and thermal cameras, each of which is impacted differently in different driving contexts. For example, GNSS signals cannot be received in a tunnel, but an IMU will still function the same as before. On the other hand, dense traffic on open roads can severely hamper camera-based localization systems due to occlusions in the camera view, while GNSS-based localization may continue to work well. The dynamic change in sensor configurations, in turn, affects how high-level AV tasks, such as localization, path planning, object detection and route planning need to adapt to the driving context at hand. Therefore, AVs cannot rely on a fixed sensor configuration or hardwired setting to handle all possible driving contexts. Rather, an AV should determine its current driving context and emphasize sensor configurations and software functions that are best suited for it.

In the following two subsections, we give examples of how various driving contexts affect different AV sensors and functions. The discussion is not meant to be exhaustive, but covers a set of common examples to highlight the need for context awareness in AVs.

| Context Description | Most Relevant AV Subsystems |
| --- | --- |
| Daytime | Perception |
| Night-time | Perception |
| Twilight | Perception |
| Sunny | Perception |
| Rainy | Perception |
| Snowy | Perception |
| Foggy | Perception |
| Dust/Sandstorm | Perception |
| Trees Overhead | Localization |
| Paved Road | Planning, Behavior |
| Lane Markers Visible | Planning, Behavior, Localization |
| Off Road | Controls, Behavior |
| Parking lot | Behavior, Perception, Localization |
| Indoors | Planning, Behavior, Localization |
| Outdoors | Planning, Behavior, Localization |
| Tunnel | Localization, Behavior |
| Urban Canyon | Localization |
| Rural area | Planning, Behavior |
| City | Planning, Behavior |
| Highway | Planning, Behavior |
| Construction Zone | Planning, Behavior, Localization |
| Heavy Traffic | Planning, Behavior |
| Bridge | Localization, Perception |
| Underpass | Localization, Perception |

TABLE I
CONTEXTS AND THEIR RELEVANCE IN AV SUBSYSTEMS

### B. Effects of Operating Conditions

We list below some dominant conditions for AVs and how they impact driving performance.

- **Weather:** Bad weather such as rain, snow, fog and duststorms severely impact camera images on an AV. Hence, an AV cannot rely solely on camera-based localization or object detection under such conditions. It can instead rely on alternative sensory inputs such as radars that are not affected by weather. A GNSS receiver may still be usable for localization and path planning.
- **Lighting:** Poor lighting conditions affect cameras, but sensory inputs such as lidar, radar, GNSS and IMUs remain unaffected. In such cases, an AV cannot rely on camera-based perception alone, but would instead need to fuse other sensors such as lidar and radar for object detection and tracking.
- **Lanemarker Visibility:** Faded lanemarkers severely hamper vision-based lane detection systems. In this situation, an AV may have to rely on maps to obtain lane information. It may also need highly accurate lidar/IMU/GNSS-based subsystems to localize itself accurately on the pre-defined maps.
- **Dense Traffic:** Some localization techniques rely on vision-based understanding of road edges and nearby stationary features [15]. Dense traffic around an ego vehicle can lead to occlusions in the camera and lidar views, making camera and lidar-based localization difficult. In this case, GNSS signals and a pre-loaded map can help the AV drive safely.

### C. Impact of Operating Environments

Next, we discuss the impact of driving contexts on the performance of AV subsystems.

- **Urban Canyon:** GNSS signals cannot be received from a sufficient number of satellites and can also induce large multi-path errors in urban canyons. As a result, an AV may need to resort to camera/lidar-based localization, path planning and navigation.
- **Tunnels:** GNSS signals are completely blocked off inside a tunnel, thus removing a source of absolute location information. Tunnels are also often visually uniform throughout and lack distinct visual features. In such cases, an AV may need to use lane matching and IMU-based dead reckoning for localization.
- **Work Zones:** Work zones pose very challenging situations for AVs due to the presence of multiple workzone objects (e.g., cones, barrels, vertical panels, barriers and signs), blocked lanes and stationary/moving vehicles such as repair and maintenance trucks. Specialized object detection techniques that fuse camera and lidar may, therefore, be required for workzone boundary detection [1]. AVs may also be able to rely on V2X infrastructure and cloud-based instructions for acquiring relevant information about the workzone.

### D. Our New *DrivingContexts* Dataset for Context Recognition in AVs

Table I presents a list of driving contexts we target for our AV dataset. We have identified 24 such contexts with binary values for each entity, resulting in a total of $2^{24}$ or over 16 million combinations, a rather large number for generating all possible combinations of training data. These contexts have been chosen based on prior literature on environmental contexts affecting AV tasks. We choose contexts in a way such that one context might exist without another, although the presence of one might make others more (or less) likely. For instance, day-time in an urban canyon might likely be correlated with heavy traffic. To be able to train a neural network to reliably detect each combination of conditions in the real world, an annotated image-based dataset should have all such combinations with multiple instances of each type. However, most publicly available datasets lack such driving context annotations and any relevant scene metadata is relatively unstructured [3]. Owing to these constraints, fully supervised learning leveraging context annotations is not readily feasible. Towards this end, we propose zero-shot and few-shot approaches in Section III-E.

The dataset that we create is named *DrivingContexts*. It comprises two subsets: *DrivingContexts* (Hand-Annotated or HA) and *DrivingContexts* (Machine-Annotated or MA). The hand-annotated subset contains images taken from a front-facing camera mounted on a vehicle and is partially sourced from popular AV datasets such as KITTI [4] and NuScenes [3]. Additionally, we also use images recorded by driving the CMU AV in and around the city of Pittsburgh, USA. To ensure that all of the relevant contexts have some scenarios for annotation, we also web-crawled copyright-free images on the Web taken from a front-facing camera for these driving contexts. We annotated by hand every image in the dataset with the context classifications listed in Table I. This hand-annotated (HA) part of the dataset comprises 1467 images across a variety of contexts along with 1467 × 24 annotations. Specifically, out of these, 500 images are from the KITTI dataset, 300 are from NuScenes, 321 are taken in Pittsburgh and the remaining 346 are from the Web. The subset from the Web was necessary because datasets such as NuScenes do not have any instances for some of the context categories such as snow, tunnel and off-road. Each image in the dataset is associated with 24 context queries hand-annotated resulting in the total of 35,208 image-query pairs.

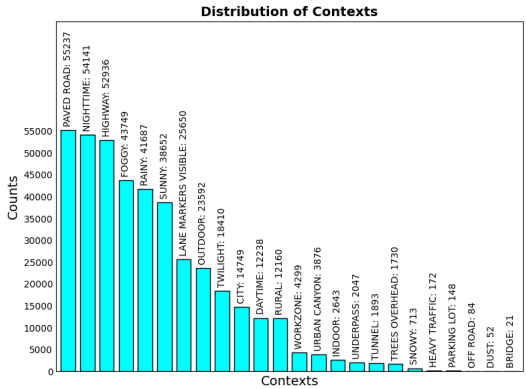

Fig. 2. Distribution of contexts in the *DrivingContexts* dataset

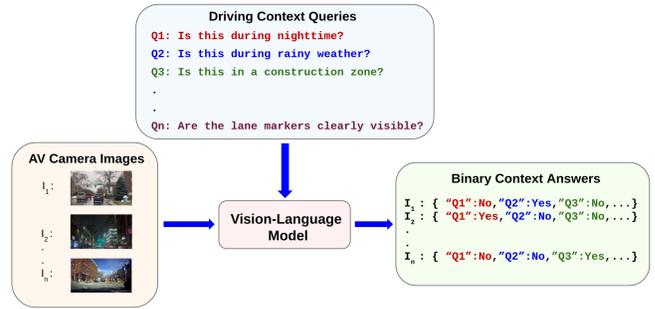

Fig. 3. Overview of *ContextVLM*

Since the number of images in the *DrivingContexts (HA)* subset is relatively small due to the constraints of manual annotation, we augment it with the much larger *DrivingContexts (MA)* subset to ensure its usefulness for evaluating *Context Recognition* approaches.

The *DrivingContexts (MA)* subset consists of 66,647 driving images. These images were recorded using a front-facing camera on our CMU AV. To capture a wide range of contexts, the CMU AV was driven in and around Pittsburgh and State College, Pennsylvania, USA, in addition to the ∼215 km route connecting these cities. A large subset of this data was collected at night-time, foggy and rainy conditions, resulting in a more challenging dataset for perception tasks. With 24 contexts per image, we have a total of about 1.6 million image-context pairs. Since this is a prohibitively large number to annotate by hand, we used a combination of multiple pre-trained general-purpose vision-language models to perform the annotation. Specifically, a context label is added to our dataset only if it has been predicted with more than 90% confidence by all of the VLMs. Furthermore, a sizeable sample of this subset was subsequently manually verified to ensure that the VLM-annotated contexts are appropriate. More details about these VLMs are provided in the following sections. We create our *DrivingContexts* dataset with context queries and corresponding annotations in the standardized VQA v2 dataset format [16] for ease of use with various VLMs. The distribution of specific contexts in our *DrivingContexts* dataset is shown in Figure 2. It can be observed that typical contexts such as highway, paved road and sunny are quite frequent. In addition, images with adverse driving conditions such as night-time, rainy and foggy attributes make the dataset useful. However, contexts such as indoors, tunnels, bridges and off-road conditions are fewer in number. This happens due to the natural, long-tailed distribution in real-world datasets, which is often a common yet complex problem in AVs and also corresponds well with Zipf's Law in language modeling. However, as we will see in Section IV, vision-language models still help significantly in accurately identifying these contexts.

### E. Our Proposed VLM-based Approach

To leverage our *DrivingContexts* dataset, we propose the use of vision-language models for detecting contexts to counter difficulties in traditional supervised learning due to the lack of adequate datasets. These VLM models were trained with image-text pairs obtained from datasets such LAION-400M and Contextual-12M that use web-crawling to obtain data and show strong zero-shot generalizability. These datasets contain image-text pairs of the order of millions which helps with multimodal representation learning and image-to-text alignment. This helps in generalization to complex tasks such as context detection without fine-tuning on domain-specific datasets. Notably, logically implausible combinations such as sandstorms and snow existing together or lane markers being visible in an off-road setting do not need to be explicitly modeled.

Figure 3 shows an overview of our *ContextVLM* approach. We take camera images generated by an AV along with a set of *n* operating context queries, where *n* represents the number of relevant contexts of interest. These are fed as inputs to a multimodal vision-language model that tokenizes the operating context query and extracts relevant features from the image. Next, the vision-language model reasons about the context query to identify relevant text such as `nighttime` from Q1, as shown in the figure. The VLM attempts to find corresponding parts of the image relevant to this question. In this specific case for detecting night-time operation, it is likely that the top part of the image which represents the sky influences the decision. Based on the appearance of relevant parts of the image, the VLM draws a conclusion on whether the context query should yield a "Yes" or "No" answer. The same strategy is used for all context queries to create a list of relevant contexts for *context recognition*.

In our work, we use two competing models as the VLM in Fig. 3, namely ViLT [17] and LLaVA [18]. These are both state-of-the-art models for Visual Question Answering (VQA) with different complexities and model sizes. ViLT is a relatively lightweight model with 87.4 million parameters, while LLaVa is a large VLM with 13 billion parameters.

*1) ViLT:* We use the Vision-and-Language Transformer (ViLT) model as one of our VLMs. It uses a Vision-and-Language pre-training paradigm that does not use deep convolutional encoders, which typically consume a large fraction of time for processing images. ViLT divides input images into patches and trains specifically for the task of image-text matching. It uses transformers for embedding vision and text modalities in a similar fashion, ensuring that patches from each image (such as "sky", "tunnel" or "lane markers" in the AV context) are associated with corresponding textual descriptions. ViLT is pre-trained on a combination of the MS COCO, Visual Genome (VG), Google Conceptual Captions (GCC) and SBU Captions datasets and the version we use is finetuned on the VQA v2 dataset [16]. With more than 10M image-text pairs from all these contexts, ViLT exhibits strong zero-shot and few-shot generalizability.

*2) LLaVa:* VLMs such as GPT-4 [19] are a larger class of models that are able to answer questions based on images. Large Language-and-Vision Assistant (LLaVA) is an open-source alternative to GPT-4 that shows comparable performance for visual understanding and VQA tasks. LLaVa has a vision and a language encoder where Vicuna [20] is used for the language component while Contrastive Language-Image Pretraining (CLIP) [21] is used for the vision component. Since LLaVA is a much larger model but requires more inference time, we will also compare the usefulness of this approach for context understanding.

## IV. EXPERIMENTAL RESULTS

In this section, we present results obtained from several experiments to demonstrate the usefulness of our dataset and model. *DrivingContexts* has varying image sizes ranging from low-resolution images of size $100 \times 100$ pixels obtained from the Web to high-resolution images generated by the CMU AV of size $1920 \times 1080$. We show that *ContextVLM* is capable of handling these varying sizes effectively and produces consistent context predictions.

### A. Quantitative Evaluation

First, we present some quantitative experimental results showing the capability of *ContextVLM* in various settings.

*1) **Zero-shot Evaluation**:* Our task involves binary classification for each context category from Table I and a generative VLM-based approach is used to accomplish this objective. The text prompts are supplied in the format described in Figure 3. We first evaluate the performance of *ContextVLM* on the smaller *DrivingContexts (HA)* dataset and its appropriate subsets. The values of accuracy, precision, recall and F1 score are reported for each subcategory in Table II. It is evident from the table that, between the two

| Method | Acc. | Precision | Recall | F1 Score |
|---|---|---|---|---|
| **DrivingContexts (Overall)** | | | | |
| ContextVLM:LLaVA | 90.66 | 82.18 | 82.03 | 82.11 |
| ContextVLM:ViLT | **93.03** | **83.56** | **91.31** | **87.26** |
| **Kitti Subset** | | | | |
| ContextVLM:LLaVA | 92.39 | 83.15 | 86.14 | 84.62 |
| ContextVLM:ViLT | **96.69** | **92.08** | **94.5** | **93.28** |
| **NuScenes Subset** | | | | |
| ContextVLM:LLaVA | 91.54 | **83.86** | 82.24 | 83.04 |
| ContextVLM:ViLT | **92.38** | 80.18 | **92.6** | **85.95** |
| **Pittsburgh Subset** | | | | |
| ContextVLM:LLaVA | 90.36 | **85.69** | 81.34 | 83.46 |
| ContextVLM:ViLT | **91.10** | 83.91 | **86.94** | **85.40** |
| **Web Subset** | | | | |
| ContextVLM:LLaVA | 87.66 | 75.99 | 77.08 | 76.53 |
| ContextVLM:ViLT | **90.10** | **76.09** | **90.54** | **82.69** |

TABLE II
ZERO-SHOT PERFORMANCE (IN %) OF *ContextVLM* ON *DrivingContexts (HA)* AND ITS SUBSETS

VLMs we consider, *ContextVLM:ViLT* performs better in most cases. The accuracy obtained is about 93% on the whole dataset, suggesting that such an approach of using VLMs for detecting contexts works reasonably well on varied real-world data. Moreover, across the subsets, KITTI is the easiest to classify well for both the *ContextVLM:ViLT* and *ContextVLM:LLaVA* approaches, also obtaining an accuracy of over 96% with precision and recall values over 90%. We note that context recognition for Web-based data is harder owing to the data being more irregular and diverse in nature in addition to having lower resolutions. Moreover, since we have identified that using ViLT as the VLM backbone works significantly better, we focus on ContextVLM with ViLT for further experiments.

Figure 4 shows the performance of the better-performing model *ContextVLM:ViLT* across all the 24 contexts from Table I in the *DrivingContexts (HA)* dataset. Our zero-shot approach performs really well with over 98% accuracy even on classes such as "tunnel" and "bridge" which are quite rare in the dataset and would therefore be difficult for supervised approaches to predict. Lane markers are relatively more difficult to classify accurately, owing to other markers on the road such as crosswalks, which played a role in increasing false detections. We expect that "prompt" engineering techniques and auto-prompt learning strategies such as Co-op [22] can aid towards further improving these predictions.

*2) **Zero-shot vs Few-shot - An Ablation Study**:* Motivated by zero-shot experiments of multiple models on the *DrivingContexts (HA)* subset, we now focus on few-shot learning. This specifically refers to experiments with supervised fine-tuning with a handful of annotated examples in a new domain. Now, for this part of our experiments, we use both the HA and MA subsets of *DrivingContexts* for an extensive evaluation. In Figure 5, we evaluate the few-shot capability

**Accuracy as a fraction across 1000 instances per context and Average Confidence for "Yes" with *ContextVLM***

| Question | Boreas[23] | | | AWD[24] |
|---|---|---|---|---|
| | Day, Sunny | Day, Snowy | Day/Dusk, Rainy | Night, Rainy/Snowy |
| "Is this during daytime?" | 1.000, 0.999 | 0.998, 0.995 | 0.413, 0.955 | 1.000, 0.000 |
| "Is this during nighttime?" | 1.000, 0.000 | 0.998, 0.002 | 0.626, 0.925 | 1.000, 0.991 |
| "Is this during sunny weather?" | 1.000, 0.999 | 1.000, 0.000 | 0.985, 0.012 | 1.000, 0.002 |
| "Is this during rainy weather?" | 1.000, 0.000 | 0.756, 0.947 | 0.922, 0.990 | 0.691, 0.918 |
| "Is this during snowy weather?" | 1.000, 0.000 | 1.000, 0.999 | 1.000, 0.000 | 0.764, 0.919 |

TABLE III
EVALUATION OF *ContextVLM* WITH ViLT ON ADVERSE WEATHER AND LOW-LIGHTING DATA FROM BOREAS [23] AND AWD [24]

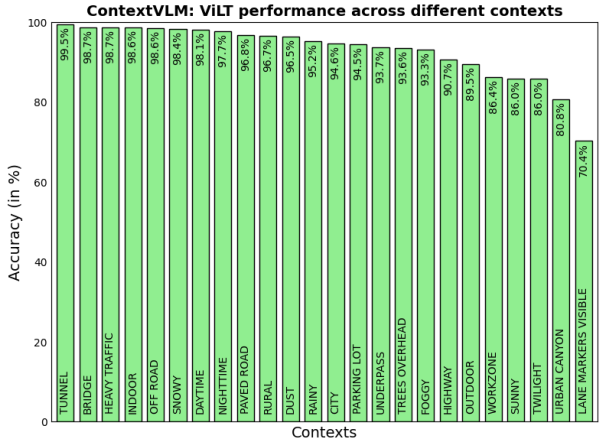

Fig. 4. Accuracies across different driving contexts for *ContextVLM:ViLT*

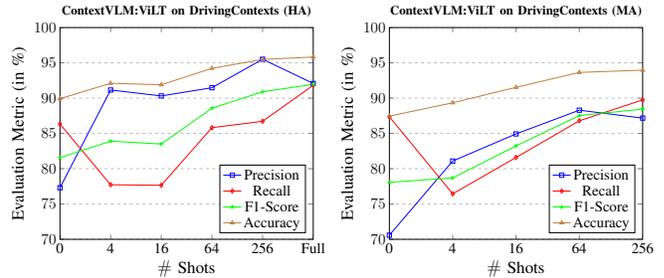

Fig. 5. Zero and Few-shot performance comparison of *ContextVLM:ViLT* on the HA and MA subsets of our *DrivingContexts* dataset

of the better-performing *ContextVLM:ViLT* model on unseen data with precision, recall, F1-score and accuracy as our metrics. Specifically, we use the pretrained model used for zero-shot evaluation and fine-tune the model with 4, 16, 64 and 256 shots from the two subsets of *DrivingContexts*.

We perform a 70:30 split on our data for training and testing respectively. For training, we use a batch size of 256 on an Nvidia A6000 GPU and evaluate few-shot performance by fine-tuning for 10 epochs for each experiment similar to settings described by Kim et al. [17]. In Figure 5, it is seen that strong performance indicators such as accuracy and F1-Score significantly improve on using our dataset in comparison to pre-existing models with zero-shot inference. Recall initially tends to drop with training using very few shots owing to the fine-tuned model being more cautious in its positive predictions with limited data to learn from, but performs well with 64 or more shots. Notably, using only 256 examples for few-shot training on both the HA and MA subsets of our *DrivingContexts* dataset improves F1-Score by around 10% and accuracy by over 5%. For the smaller HA subset of our *DrivingContexts* dataset, we also fine-tune on the whole training subset for comparison with few-shot approaches. From Figure 5, we also notice that in general, most of the improvement in performance by fine-tuning is obtained just by using about 64 examples for few-shot learning.

In a broader context, similar to observations made for the capabilities of language models [25], *ContextVLM:ViLT* generalizes well with a small number of examples to a new *Context Recognition* task. In comparison to traditional supervised computer vision models which require large an-notated datasets, our few-shot approach with VLMs is useful for generalization across new driving contexts with a small number of annotated examples.

*3) Evaluation on Public Video Streams:* In addition to our annotated dataset *DrivingContexts*, we also evaluate *ContextVLM* on publicly available adverse weather video streams. Specifically, we use videos from two public AV datasets, Boreas [23] and the Adverse Weather Dataset (AWD) [24], whose images are predominantly from snowy, rainy and night-time conditions. Hence, we extract frames from videos and evaluate our approach on a total of 4,000 *additional* annotated images. For this subset of data, the performance of *ContextVLM* (with ViLT as the VLM backbone) is presented in Table III. The number of correctly classified examples along with the average confidence for "yes" or "no" responses is highlighted. We see that, in adverse real-world datasets, *ContextVLM* successfully identifies the driving contexts, with an accuracy close to 100% for many driving contexts. Furthermore, our approach has over 90% confidence for each of its definitive positive answers along with lower than 0.2% average confidence when predicting incorrect answers. *ContextVLM* can be readily integrated into AV perception systems such as Hydrafusion [8] that rely on the knowledge of weather and lighting conditions.

*4) Inference Time:* Table IV shows our evaluation of the inference time per query and the total inference time for all of the driving contexts of interest on a lightweight Nvidia RTX A4000 laptop GPU. Additionally, for the LLaVA model which is generalized and can handle more complex inputs, we adopt two approaches. First, *LLaVA (Individual)* takes each question one at a time and answers with a simple *yes/no* response. Secondly, *LLaVA (Joint)* fuses all context queries together and generates a combined answer. It is observed from the first column that the joint approach has a lower inference time per question (1.141s vs 1.766s). Hence, fusing context queries can significantly reduce the timing

| VLM backbone for *ContextVLM* | Inference Time per Query (s) | Total Inference (s) |
|---|---|---|
| **LLaVA (Individual)** | 1.766 | 42.382 |
| **LLaVA (Joint)** | 1.141 | 27.470 |
| **ViLT (Individual)** | 0.039 | 0.942 |

TABLE IV

INFERENCE TIME COMPARISON ON AN NVIDIA RTX A4000 GPU

complexity of our problem. However, we also observe that the detection of contexts like lighting which may change quickly can only be done every few seconds with this approach, which may be a limitation. On the other hand, ViLT is a much simpler model and is therefore unable to handle more complex queries with multiple sub-questions. Hence, we only evaluate the *Individual* option with ViLT. ViLT has an inference time of 39 ms per query, which is more than 28 times faster than that for *LLaVA (Joint)*. This is to be expected since ViLT is a much smaller model in terms of its parameter count. However, Table II also shows that ViLT performs better and is a feasible option. In short, while traditional large Vision Language Models might be too slow for real-time inferencing on an AV, more lightweight models such as ViLT are capable of performing accurate predictions while running in real-time. Additionally, this approach has minimal GPU constraints, using less than 1GB of GPU memory at runtime.

Furthermore, although we have 24 questions per image for our generalized task, AVs operating in restricted ODDs may only be interested in a strict subset of these contexts for integration into their downstream tasks. Moreover, contexts such as lighting and weather conditions will typically not change at sub-second intervals. Later, in Section IV-C, we will discuss the timing and resource constraints for deployment on an AV.

*B. Visual Analysis*

For illustrative purposes, we now evaluate *ContextVLM:ViLT* on two images across day and night-time. For the daytime image shown on the left of Fig. 6 obtained from the CMU AV, which is also a part of *DrivingContexts*, the relevant contexts identified by *ContextVLM* are as follows: `"OUTDOOR"`, `"DAYTIME"`, `"SUNNY"`, `"HIGHWAY"`, `"RURAL"`, `"WORKZONE"`, `"PAVED ROAD"` and `"LANE MARKERS VISIBLE"`. From the image, we see that the car is being driven on a sunny day on the highway with a rural backdrop, lanes are demarcated by lane markers, and there are cones and warning signs signifying a work zone. This response can, for instance, trigger dynamic perception and behavioral computational adaptations for workzones, such as the ones discussed by Shi et al. [1] and Sural et al. [26].

For the night-time image on the right side of Fig. 6, our approach generates `"NIGHTTIME"`, `"RAINY"`, `"FOGGY"`, `"HIGHWAY"` and `"PAVED ROAD"` as the predicted contexts. The image clearly is of driving at night in the rain, which is validated by water droplets on the windshield and reflections of traffic lights on the wet road. The context recognition output generated in this case can serve as an input for multimodal object detection models such as ContextualFusion [9], which can assign higher importance to lidar and/or radar data over camera at night-time.

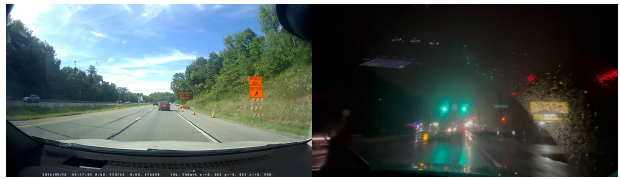

Fig. 6. Sample images from the Pittsburgh subset of *DrivingContexts* for evaluating context detection performance

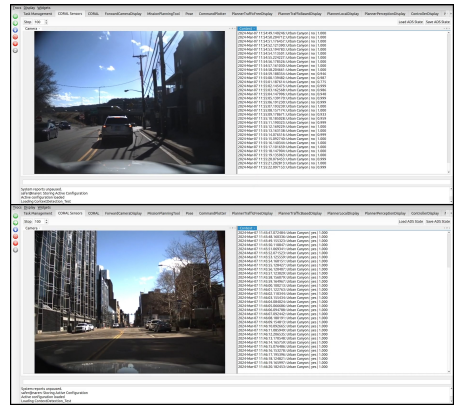

Fig. 7. ContextVLM running on the CMU AV inferring necessary contexts

*C. Edge Inference on an AV*

Based on the qualitative and quantitative results presented in the last two subsections, we conclude that *ContextVLM* with ViLT works the best overall with a high accuracy and has a relatively low overall inference time. Motivated by its practical feasibility, we implemented ContextVLM as a perception task on our CMU AV for real-time inferencing, using a front-facing camera with a resolution of $2048 \times 1536$. The images are sent to all perception tasks being run by the AV including ours for context detection. As a demonstration, we specifically asked the question- "Are there tall buildings around?" to observe the performance of *ContextVLM* in urban regions with skyscrapers.

We evaluated the response to our query by driving around in the city of Pittsburgh, Pennsylvania, USA. The upper half of Figure 7 shows the AV driving in a residential neighborhood with no tall buildings. In this case, *ContextVLM* answers negatively to the presence of urban canyons with a very high confidence ($> 99\%$). In the lower half of the image, there are skyscrapers to the left in downtown Pittsburgh and *ContextVLM* correctly predicts the presence of an urban canyon.

This context query was chosen specifically to aid in localization tasks that rely on GNSS and other input sources such as cameras. The presence of urban canyons can significantly affect the performance of GNSS sensors and other localization algorithms need to come into play at this point. Reliable *ContextVLM*-based outputs have a significant positive impact in this case while running in real-time with low latency. Running this task on a lightweight 4GB Nvidia GeForce

GTX 1050 Ti GPU on the CMU AV results in a latency of 10.5 ms per query. In the worst case, even if all 24 queries need to be posed, context recognition can occur at a rate of 4 Hz. We expect much better performance in future AV hardware. For other perception tasks such as those that need information about rain to perform well, *ContextVLM* can be applied in its current form by only changing the context query without any further training needed.

## V. CONCLUSIONS

Widespread deployment of AVs requires that they operate in a wide range of driving contexts including different weather, lighting, traffic and road conditions. In this paper, we have proposed an approach called *ContextVLM* that uses a vision-language model for identifying the environmental and driving contexts of an AV. This approach deviates from traditional fully supervised methods for detecting specific driving contexts such as rain and night-time, which do not scale well to a large number of contexts. We also created a large dataset called *DrivingContexts* with more than 68,000 images with a total of 1.63 million annotations.

Our *ContextVLM* model produces an accuracy of over $95\%$ on unseen data with few-shot training on the overall *DrivingContexts* dataset, validating the usability of our approach. Moreover, as few as 64 examples from our dataset were needed to significantly improve F1-score by over $10\%$ on unseen data, showing the applicability of our work to new driving contexts. Additionally, we achieve a low latency of 10.5 ms per query, enabling us to run *ContextVLM* in real-time on a GPU on our CMU AV. We make our dataset and code publicly available for further research in this direction. In the future, our work can be extended to incorporate common-sense reasoning methods that generate rationales to improve the understanding of driving contexts. The approach also needs to be validated extensively in real-time across various operating conditions.

## ACKNOWLEDGEMENT

This work is funded by the US Department of Transportation under its ADS program.